\documentclass{article}

\PassOptionsToPackage{numbers}{natbib}
\usepackage{natbib}
\usepackage[utf8]{inputenc} 
\usepackage[T1]{fontenc}    
\usepackage{hyperref}       
\usepackage{url}            
\usepackage{booktabs}       
\usepackage{amsfonts}       
\usepackage{nicefrac}       
\usepackage{microtype}      
\usepackage{xcolor}         
\usepackage{geometry}
\usepackage{soul}

\definecolor{hlgreen}{HTML}{defde0}
\definecolor{hlorange}{HTML}{fcf7de}
\definecolor{italian}{HTML}{ffcfd2}
\definecolor{german}{HTML}{f1c0e8}
\definecolor{french}{HTML}{baf2d8}
\definecolor{spanish}{HTML}{90dbf4}

\newcommand{\et}[1]{\sethlcolor{hlgreen}\hl{#1}}
\newcommand{\ot}[1]{\sethlcolor{hlorange}\hl{#1}}
\newcommand{\ita}[1]{\sethlcolor{italian}\hl{#1}}
\newcommand{\fra}[1]{\sethlcolor{french}\hl{#1}}
\newcommand{\spa}[1]{\sethlcolor{spanish}\hl{#1}}
\newcommand{\ger}[1]{\sethlcolor{german}\hl{#1}}

\usepackage{microtype}
\usepackage{graphicx}
\usepackage[caption = false]{subfig}
\usepackage{float}
\title{Adversarial Attacks on Image Generation\\With Made-Up Words}
\date{}

\author{%
  Rapha{\"e}l Milli{\`e}re \\
  Columbia University  \\
  \texttt{rm3799@columbia.edu}
}

\begin{document}

\maketitle

\begin{abstract}
Text-guided image generation models can be prompted to generate images using made-up words adversarially designed to robustly evoke specific visual concepts. Two approaches for such generation are introduced: \emph{macaronic prompting}, which involves designing cryptic hybrid words by concatenating subword units from different languages; and \emph{evocative prompting}, which involves designing nonce words whose broad morphological features are similar enough to that of existing words to trigger robust visual associations. The two methods can also be combined to generate images associated with more specific visual concepts. The implications of these techniques for the circumvention of existing approaches to content moderation, and particularly the generation of offensive or harmful images, are discussed.
\end{abstract}

\section{Introduction}

Text-guided image generation models have made impressive strides in recent years. State-of-the-art models, like DALL-E 2~\citep{rameshHierarchicalTextConditionalImage2022}, Imagen~\citep{sahariaPhotorealisticTexttoImageDiffusion2022}, and Parti~\citep{yuScalingAutoregressiveModels2022}, can generate coherent images matching a remarkably wide variety of prompts in virtually any visual domain and style. While the ability to generate high-quality images of any subject is an exciting development for content creation, it also raises ethical questions about potential misuse of this technology. In particular, text-guided image generation models may be used to produce fake imagery of existing individuals for misinformation (so-called ``deepfakes'' \citep{milliereDeepLearningSynthetic2022}), or produce visual content deemed offensive or harmful. These concerns have been used to justify the decision to limit access to large text-guided image generation models, as well as moderate their use according to content policies implemented in prompt filters.

One particular concern regarding the robustness and apppropriate use of deep neural networks is the existence of so-called adversarial examples designed to mislead them. Adversarial examples have been extensively investigated in computer vision, where adding a small but carefully crafted perturbation to an image can lead image classifiers, but not humans, to make radical classification errors \citep{szegedyIntriguingPropertiesNeural2014,goodfellowExplainingHarnessingAdversarial2015}. The process of generating adversarial examples that can fool a given neural network is also known as an adversarial attack. In recent years, such attacks have been investigated beyond image classifiers. While adversarial perturbations can be indistinguable to the original input for humans in the visual domain, this is not typically the case in the linguistic domain, because text data is discrete rather than continuous like pixel values (although see \citep{boucherBadCharactersImperceptible2022} for an example of hidden adversarial attacks on text). Nonetheless, various alternative approaches for adversarial attacks have explored in natural language processing (NLP), causing models to misclassify text or generate biased or harmful text with minimal perturbations of the input \citep{zhangAdversarialAttacksDeeplearning2020}. For example, appending a seemingly meaningless sequence of tokens to a paragraph can cause NLP models to fail at question answering tasks or spew racist outputs \citep{jiaAdversarialExamplesEvaluating2017,wallaceUniversalAdversarialTriggers2021}. 

Adversarial examples have also been explored with vision-language models designed for image captioning and recognition \citep{radfordLearningTransferableVisual2021}. For example, so-called \emph{typographic attacks} involve applying a real-life erroneous label to an item in an image (such as sticking a piece of paper with ``ipod'' written on it on an apple), causing vision-language models to misclassify that item \citep{goh2021multimodal}. However, adversarial attacks on text-guided image generation models have not been thoroughly investigated to date, despite common concerns about their potential misuse. Recent work suggests that it is possible to generate images corresponding to specific visual concepts using seemingly nonsensical prompts with some text-guided image generation models. Thus, Daras \& Dimakis~\citep{darasDiscoveringHiddenVocabulary2022} suggest that gibberish text depicted in images generated by DALL-E 2 may belong to a ``hidden vocabulary'', which can be transcribed and used in prompts to generate specific kinds of images. For example, they observed that the string \texttt{Apoploe vesrreaitais}, appearing in at least one image generated by the model, could be used as a prompt to yield images of birds. This method is potentially problematic, as it could be generalized to perform attacks on text-guided image generation models to get around content moderation filters. While this is not an adversarial attack in the traditional sense of introducing small or imperceptible perturbation to a model's input in order to drastically change its output, it does fall within the category of adversarial attacks in a broader sense: inputs are adversarially crafted to be meaningless and seemingly innocuous to humans, yet produce specific outputs associated with determinate visual concepts. If nonce strings can reliably be used to generate specific imagery with image generation models, then prompt filtering based on blacklisted words (e.g., violent, racist, sexist, or pornographic concepts) might turn out to be ineffective. 

It is currently unclear how robust the association between gibberish strings and visual concepts is in text-guided image generation models. Many of the strings depicted in images generated by DALL-E 2 do not seem consistently associated with particular visual concepts. Even those that do, like \texttt{Apoploe vesrreaitais}, may not be associated to visual concepts with the same degree of robustness as real English words like \texttt{bird}. Furthermore, this phenomenon may be partially or wholly attributable to tokenization with \emph{byte pair encoding} (BPE)~\citep{10.5555/177910.177914,sennrichNeuralMachineTranslation2016} used to train the CLIP model used for DALL-E 2~\citep{rameshHierarchicalTextConditionalImage2022}. Thus, \texttt{Apoploe vesrreaitais} is tokenized with BPE as \texttt{\et{Apo}\ot{plo}\et{e}} \texttt{\et{ve}\ot{sr}\et{re}\ot{ait}\et{ais}} (alternating colors indicate subword segmentation). Interestingly, \emph{Apodidae} and \emph{Ploceidae} are Latin names for bird families, and get tokenized by CLIP as \texttt{\et{apo}\ot{di}\et{dae}} and \texttt{\et{plo}\ot{ce}\et{ida}\ot{e}} respectively. It is possible that the association between \texttt{Apoploe vesrreaitais} and images of bird is at least partially mediated by sharing initial tokens with \texttt{Apodidae} and \texttt{Ploceidae}. The claim that DALL-E 2 has developped a ``hidden vocabulary'' may be misleading, but highlights the importance of investigating the relationship between subword tokenization and image generation in text-guided image generation models, and, more broadly, viable strategies for adversarial attacks with nonce strings.

One significant limitation of the adversarial approach proposed by Daras \& Dimakis~\citep{darasDiscoveringHiddenVocabulary2022} is that it does not offer a reliable method to find nonce strings that elicit specific imagery. Most of the gibberish text included by DALL-E 2 in images does not seem to be reliably associated with specific visual concepts when transcribed and used as a prompt. This limits the viability of this approach as way to circumvent the moderation of harmful or offensive content; as such, it is not a particularly concerning risk for the misuse of text-guided image generation models. The present work introduces two alternative methods for adversarial attacks on text-guided image generation models using nonce strings. \emph{Macaronic prompting} involves composing multilingual subword units to generate specific imagery from seemingly nonsensical strings, while \emph{evocative prompting} relies on inventing words whose broad morphological features trigger consistent semantic associations with real concepts. Both methods can potentially be used to avoid triggering content filters, which pauses a potential risk for the safe deployment of text-guided image generation models. Macaronic prompting, in particular, appears to be significantly more reliable than the method proposed by Daras \& Dimakis~\citep{darasDiscoveringHiddenVocabulary2022} to prompt such models with nonce strings. Furthermore, the allows the same adversarial attacks to work effectively on different models, instead of being tailored to one specific model like DALLE-2.

\section{Macaronic prompting}

The term ``macaronic'' traditionally refers to a mixture of several languages. One specific phenomenon that falls under this broad category is lexical hybridization, a form of code mixing in which words or morphemes from multiple languages are combined into novel words. Lexical hybridization spontaneously occurs in many multilingual communities. For example, Urdu-English code mixing is common in Pakistan, and includes the combination of English nouns with Urdu suffixes \citep{dilshadVerbHybridizationInteresting2007}.

Text-guided image generation models using subword tokenization learn robust associations between subword units and visual concepts. Furthermore, they typically learn such associations in multiple languages, as training datasets scraped from the internet often contain image-caption pairs in different languages, even when multilinguality was not explicitly part of the curation goals. The core idea of macaronic prompting is to leverage such associations across different languages, in a way that purposefully obfuscates the nature of the visual concepts targeted in the prompt. This is achieved through a form of artificial lexical hybridization that involves combining multilingual subword units associated with similar visual concepts to produce nonce strings that do not trigger robust semantic associations in humans.

\begin{figure}[!htp]
\centering 
\includegraphics[width=\textwidth]{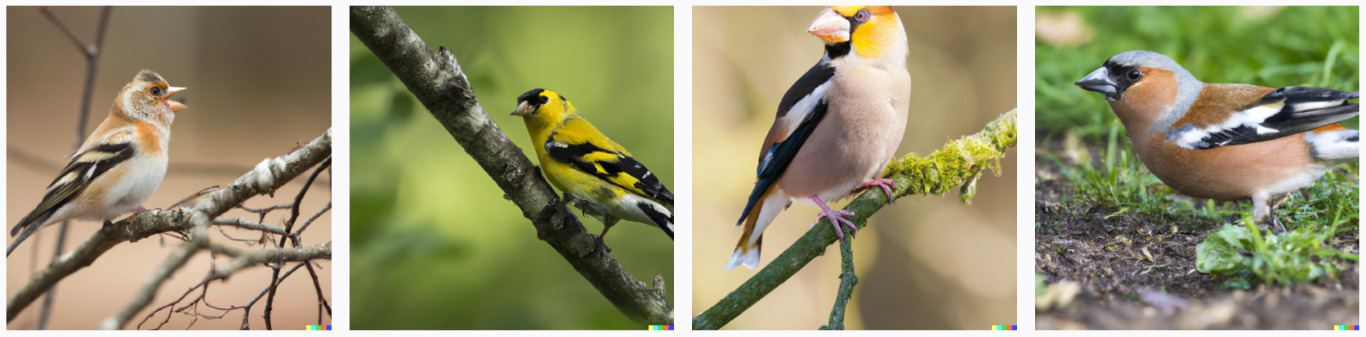}
\vspace*{-7mm}
\caption{Sample images generated by DALL-E 2 with the prompt \texttt{uccoisegeljaros}.}
\label{fig:birds1}
\end{figure}

For example, the word for ``birds'' is \texttt{Vögel} in German, \texttt{uccelli} in Italian, \texttt{oiseaux} in French, and \texttt{p{\'a}jaros} in Spanish. With the BPE tokenization method used by CLIP, these words are tokenized as follows: \texttt{\et{v}\ot{o}\et{gel}}, \texttt{\et{ucc}\ot{elli}}, \texttt{\et{o}\ot{ise}\et{aux}}, and \texttt{\et{p}\ot{a}\et{jar}\ot{os}}. Subword tokens from these words can be creatively combined to form a nonce words that looks like gibberish to human speakers, such as \texttt{\ita{ucc}\fra{oise}\ger{gel}\spa{jaros}}.\footnote{For clarity, the following color scheme is used to indicate the origin of each part of the nonce string: \ger{German}, \ita{Italian}, \fra{French}, and \spa{Spanish}.} Sample images generated by DALL-E 2 using this nonce string are shown in fig. \ref{fig:birds1}.

\begin{figure}[!htp]
\centering 
\subfloat[Images generated by DALL-E 2 with the prompt \texttt{voiscellpajaraux}.]{\includegraphics[width=0.45\textwidth]{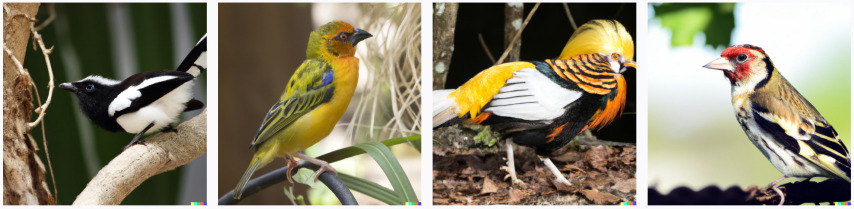}} \hspace{2mm}
\subfloat[Images generated by DALL-E 2 with the prompt \texttt{oisvogajaro}.]{\includegraphics[width=0.45\textwidth]{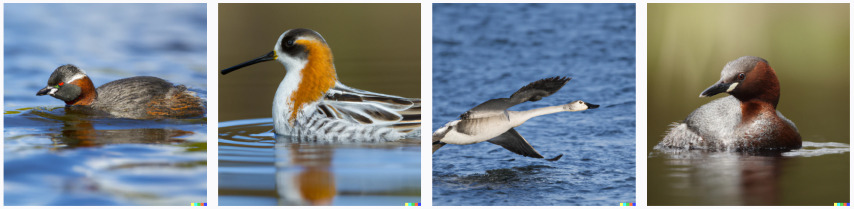}} \hspace{2mm}
\caption{Examples of adversarial macaronic prompting that do not strictly respect the boundaries of subword tokenization for lexical hybridization.}
\label{fig:birds2}
\end{figure}

Actual examples of code mixing in linguistic communities need not respect the so-called ``equivalence constraint'', according to which languages will tend be switched at points where their surface structures map onto each other \citep{poplackSometimesLlStart1980}. By analogy, macaronic prompting need not strictly respect subword segmentation driven by BPE tokenization, and good results can in fact be obtained by carving out and recomposing words with arbitrary chunks. For example, \texttt{\ger{v}\fra{ois}\ita{cell}\spa{pajar}\fra{aux}} and \texttt{\fra{ois}\ger{vog}\spa{ajaro}} work just as well as \texttt{\ita{ucc}\fra{oise}\ger{gel}\spa{jaros}} to generate images of birds (fig. \ref{fig:birds2}).

\begin{figure}[!htp]
\centering 
\subfloat[Prompt: \texttt{insekafetti}.]{\includegraphics[width=0.45\textwidth]{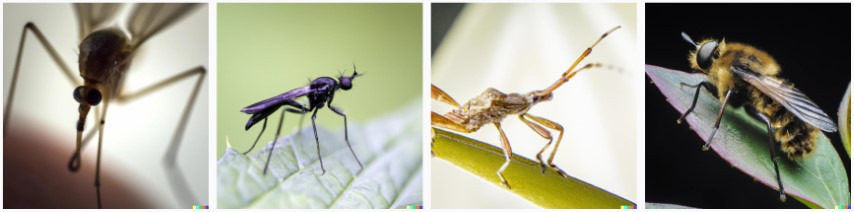}} \hspace{2mm}
\subfloat[Prompt: \texttt{farpapmaripterling}.]{\includegraphics[width=0.45\textwidth]{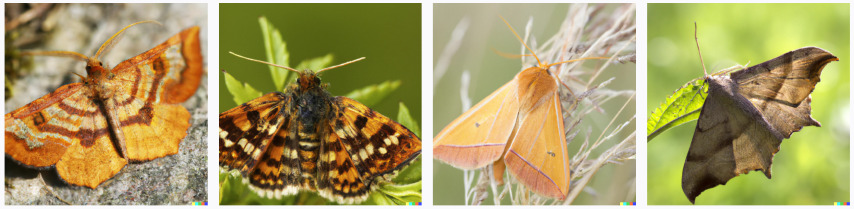}} \hspace{2mm}
\subfloat[Prompt: \texttt{eidelucertlagarzard}.]{\includegraphics[width=0.45\textwidth]{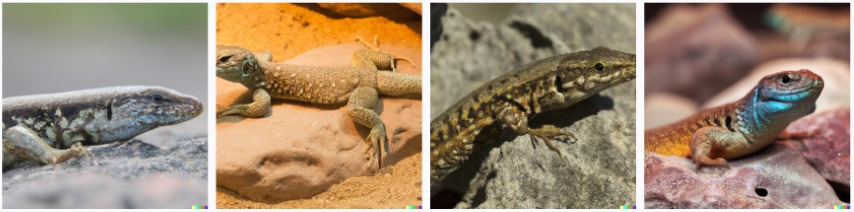}} \hspace{2mm}
\subfloat[Prompt: \texttt{coniglapkaninc}.]{\includegraphics[width=0.45\textwidth]{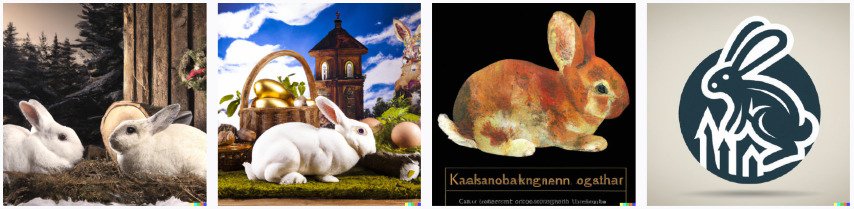}} \hspace{2mm}
\subfloat[Prompt: \texttt{falaiscoglieklippantilado}.]{\includegraphics[width=0.45\textwidth]{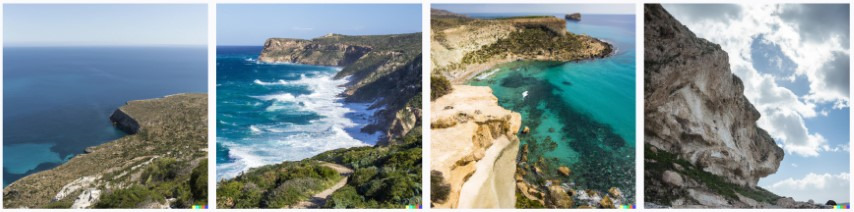}} \hspace{2mm}
\subfloat[Prompt: \texttt{avflugzereo}.]{\includegraphics[width=0.45\textwidth]{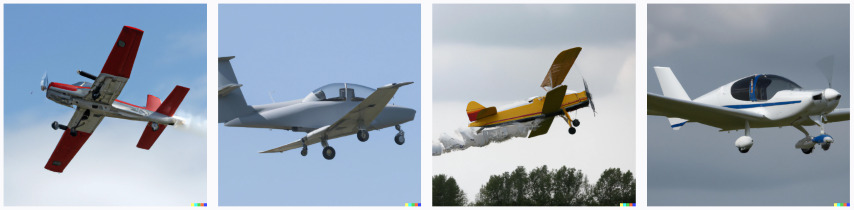}} \hspace{2mm}
\subfloat[Prompt: \texttt{feuerpompbomber}.]{\includegraphics[width=0.45\textwidth]{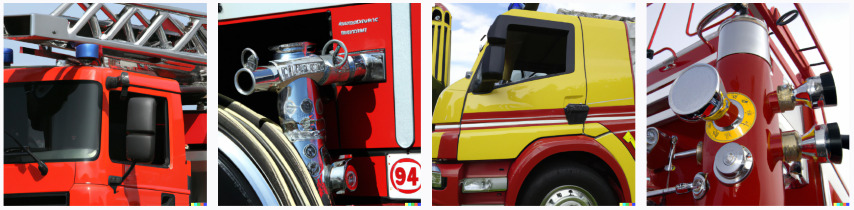}} \hspace{2mm}
\subfloat[Prompt: \texttt{educbildacion}.]{\includegraphics[width=0.45\textwidth]{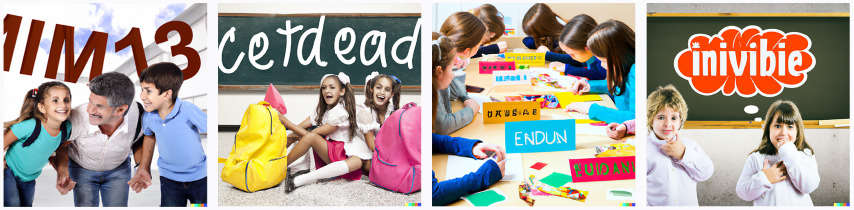}} \hspace{2mm}
\caption{Additional examples of macaronic prompting with DALL-E 2.}
\label{fig:various-dalle2}
\end{figure}

Table \ref{table:macaronic} shows additional examples of lexical hybridization used for effective macaronic prompting about various visual concepts in different domains. Non-cherry-picked sample images generated with DALL-E 2 using the resulting nonce strings are displayed in fig. \ref{fig:various-dalle2}. These samples exhibit high consistency, despite the somewhat cryptic nature of the prompts for humans.

\begin{table}[!htp]
\centering
\begin{tabular}{|c| c c c c| c|} 
 \hline
 English & German & Italian & French & Spanish & Hybridized (example) \\ 
 \hline
 bugs & \ger{K{\"a}fer} & \ita{insetti} & \fra{insectes} & \spa{bichos} & \fra{inse}\ger{kaf}\ita{etti} \\ 
 \hline
 butterfly & \ger{Schmetterling} & \ita{farfalla} & \fra{papillon} & \spa{mariposa} & \ita{far}\fra{pap}\spa{marip}\ger{terling} \\
 \hline
 lizard & \ger{Eidechse} & \ita{lucertola} & \fra{l{\'e}zard} & \spa{lagarto} & \ger{eide}\ita{lucert}\spa{lagar}\fra{zard} \\
 \hline
 rabbit & \ger{Kaninchen} & \ita{coniglio} & \fra{lapin} & \spa{conejo} & \ita{conig}\fra{lap}\ger{kaninc} \\
 \hline
 cliff & \ger{Klippe} & \ita{scogliera} & \fra{falaise} & \spa{acantilado} & \fra{falai}\ita{scoglie}\ger{klipp}\spa{antilado} \\
 \hline
 plane & \ger{Flugzeug} & \ita{aereo} & \fra{avion} & \spa{avi{\'o}n} & \fra{av}\ger{flugz}\ita{ereo} \\
 \hline
 firefighter & \ger{feuerwehrmann} & \ita{pompiere} & \fra{pompier} & \spa{bombero} & \ger{Feuer}\ita{pomp}\spa{bomber}  \\
 \hline
 education & \ger{Bildung} & \ita{educazione} & \fra{{\'e}ducation} & \spa{educaci{\'o}n} & \fra{educ}\ger{bild}\spa{acion}  \\
 \hline
 exasperation & \ger{Entt{\"a}uschung} & \ita{esasperazione} & \fra{exasp{\'e}ration} & \spa{exasperaci{\'o}n} & \fra{exasp}\ger{enttaus}\spa{acion}  \\
 \hline
\end{tabular}
\caption{Additional examples of hybridized made-up words used for macaronic prompting.}
\label{table:macaronic}
\end{table}

\begin{figure}[!htp]
\centering 
\subfloat[Images generated by DALL-E mini with the prompt: \texttt{insekafetti}.]{\includegraphics[width=0.45\textwidth]{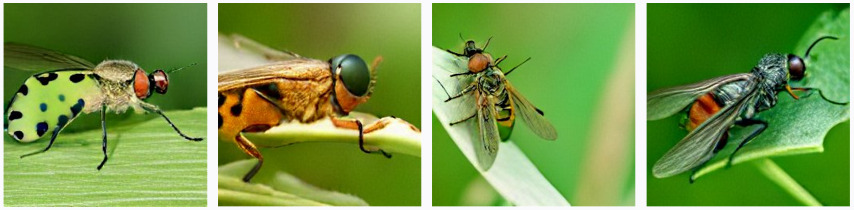}} \hspace{2mm}
\subfloat[Images generated by DALL-E mini with the prompt: \texttt{falaiscoglieklippantilado}.]{\includegraphics[width=0.45\textwidth]{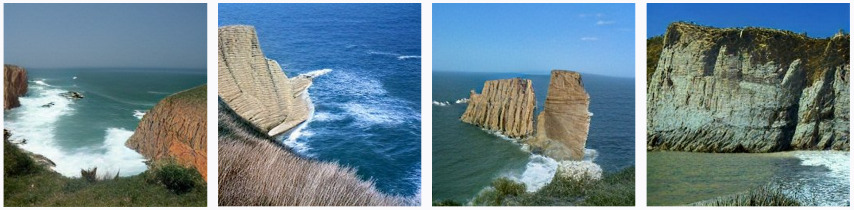}} \hspace{2mm}
\subfloat[Images generated by DALL-E mini with the prompt: \texttt{avflugzereo}.]{\includegraphics[width=0.45\textwidth]{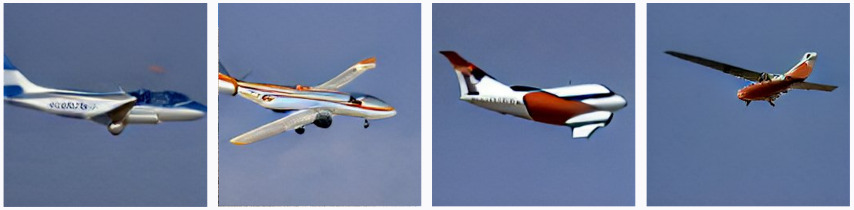}} \hspace{2mm}
\subfloat[Images generated by DALL-E mini with the prompt: \texttt{educbildacion}.]{\includegraphics[width=0.45\textwidth]{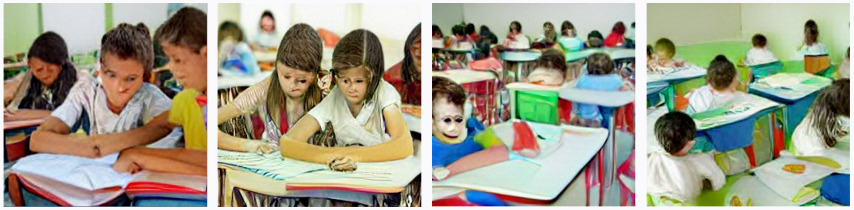}} \hspace{2mm}

\caption{Examples of macaronic prompts that work across models.}
\label{fig:across-models}
\end{figure}

While different text-guided image generation models have different architectures, training data, and tokenization methods, macaronic prompting can in principle be applied to any model trained on multilingual data. In fact, some macaronic prompts appear to work consistently across models, suggesting that it might be a viable approach for one-size-fits-all adversarial attacks on image generation algorithms. Fig. \ref{fig:across-models} shows some samples generated by DALL-E mini \citep{daymaDALLMini2021} with macaronic prompts that also work with DALL-E 2. It is worth noting that, despite their similar names, DALL-E 2 and DALL-E mini are fairly different. They have different architectures (notably, DALL-E mini does not use diffusion), are trained on different datasets, and use different tokenization procedures (DALL-E mini uses a BART tokenizer that may segment words differently than CLIP's tokenizer). Given these differences, the fact that many macaronic prompts work on both models is noteworthy.

\begin{figure}[!htp]
\centering 
\subfloat[Images generated by DALL-E mini with the prompt: \texttt{farpapmaripterling}.]{\includegraphics[width=0.45\textwidth]{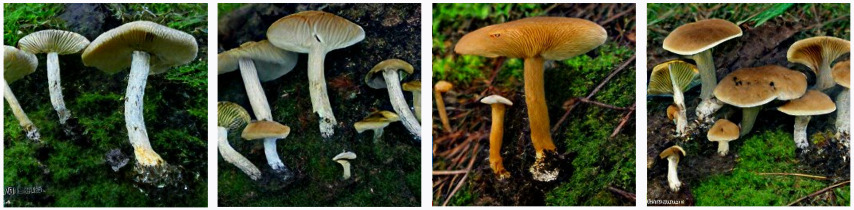}} \hspace{2mm}
\subfloat[Images generated by DALL-E mini with the prompt: \texttt{maripofarterling}.]{\includegraphics[width=0.45\textwidth]{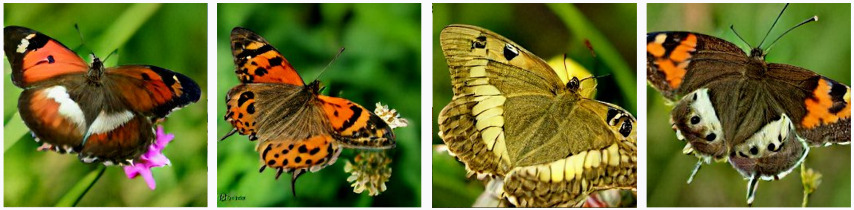}} \hspace{2mm}
\subfloat[Images generated by DALL-E mini with the prompt: \texttt{voiscellpajaraux}.]{\includegraphics[width=0.45\textwidth]{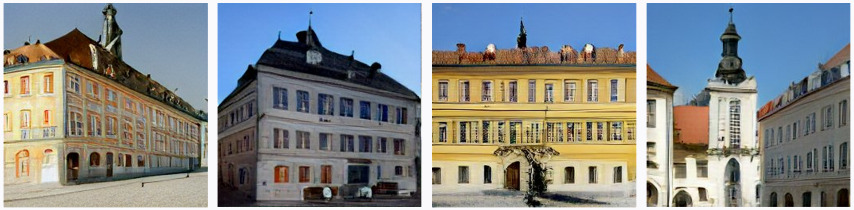}} \hspace{2mm}
\subfloat[Images generated by DALL-E mini with the prompt: \texttt{eidelucertlagarzard}.]{\includegraphics[width=0.45\textwidth]{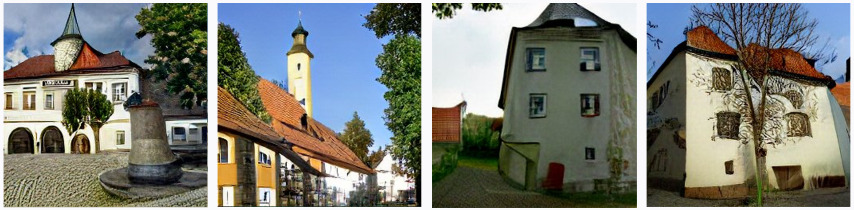}} \hspace{2mm}

\caption{Examples of transfer failure and success across models.}
\label{fig:transfer-failures}
\end{figure}

However, not all macaronic prompts transfer appropriately across models. For example, while \texttt{\ita{far}\fra{pap}\spa{marip}\ger{terling}} produces images of butterflies with DALL-E 2 as intended (fig. \ref{fig:various-dalle2}b), it unexpectedly produces images of mushrooms with DALL-E mini (fig. \ref{fig:transfer-failures}a). Interestingly, using the different hybridized nonce word \texttt{\spa{maripo}\ita{far}\ger{terling}} yields images of butterflies in both DALL-E 2 and DALL-E mini (fig. \ref{fig:transfer-failures}b). Other examples of transfer failure include \texttt{\ger{v}\fra{ois}\ita{cell}\spa{pajar}\fra{aux}} and \texttt{\ger{eide}\ita{lucert}\spa{lagar}\fra{zard}}, which produce images of houses with DALL-E mini, instead of images of birds and lizards, respectively (fig. \ref{fig:transfer-failures}c,d). It is possible that larger models trained on larger datasets are more susceptible to macaronic prompting, as they learn more robust associations between subword units and visual concepts across languages. This might explain why some macaronic prompts that yield the expected results with DALL-E 2 do not work with DALL-E mini, while the reverse does not seem to hold. This trend is potentially worrisome, as it suggests that larger models may be more vulnerable to adversarial attacks using macaronic prompting.

\begin{figure}[!htp]
\centering 
\subfloat[Images generated with the prompt: \texttt{A man in a state of exaspenttausacion}.]{\includegraphics[width=0.45\textwidth]{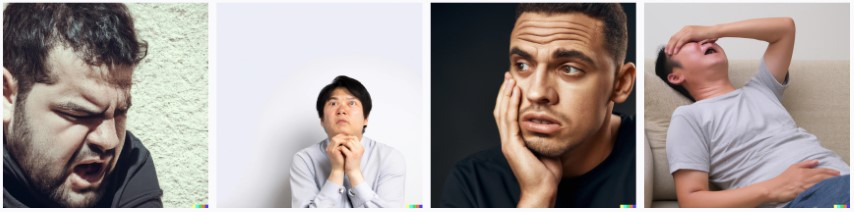}} \hspace{2mm}
\subfloat[Images generated with the prompt: \texttt{A farpapmaripterling lands on a feuerpompbomber}.]{\includegraphics[width=0.45\textwidth]{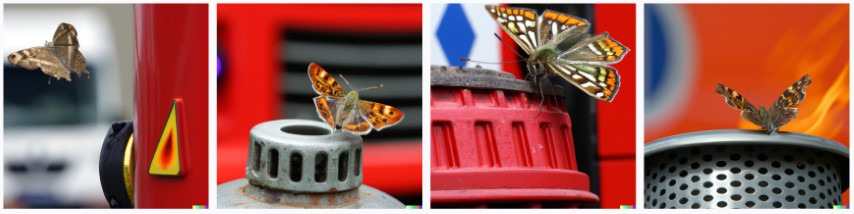}} \hspace{2mm}
\subfloat[Images generated with the prompt: \texttt{An eidelucertlagarzard maripofarterling. A fantastical hybrid creature, part eidelucertlagarzard and part maripofarterling, digital art}.]{\includegraphics[width=0.45\textwidth]{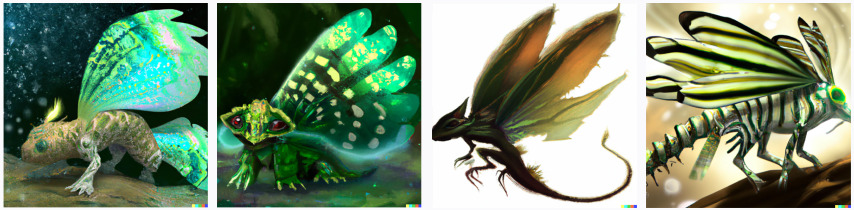}} \hspace{2mm}
\subfloat[Images generated with the prompt: \texttt{An eidelucertlagarzard eating a maripofarterling, digital art}.]{\includegraphics[width=0.45\textwidth]{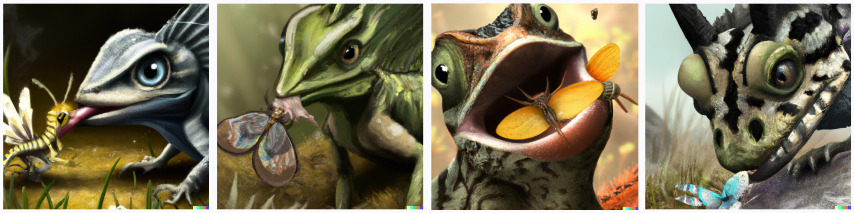}} \hspace{2mm}
\caption{Examples of compositional macaronic prompting with DALL-E 2.}
\label{fig:compositionality}
\end{figure}

In addition to their standalone use as prompts, macaronic nonce strings can be composed in sentences using English syntax, with some success. Fig. \ref{fig:compositionality} shows some examples of compositional prompts involving one or several macaronic nonce strings. The fact that such strings can be combined to generate more specific and complex scenes opens up further possibilities for the generation of potentially problematic content using this approach. While complex macaronic prompts require English (or another natural language suffuciently well-represented in the training data) as a scaffold for syntactic structure, thus making them somewhat more interpretable than prompts using unique strings, the information conveyed to the model remains relatively well obsfuscated. For example, it would probably be challenging for most people to guess what kind of scene will be generated with the prompt \texttt{An \ger{eide}\ita{lucert}\spa{lagar}\fra{zard} eating a \spa{maripo}\ita{far}\ger{terling}} without prior exposure to macaronic prompting and knowledge of the languages used for hybridization. Furthermore, compositionally complex prompts of this kind will not trigger content filters based on blacklists, despite their use of plain English words, as long as the censored concepts are adequately ``encrypted'' using the macaronic method.

\begin{figure}[!htp]
\centering 
\subfloat[Images generated by DALL-E 2 with the prompt: \texttt{A very creepooky person}.]{\includegraphics[width=0.45\textwidth]{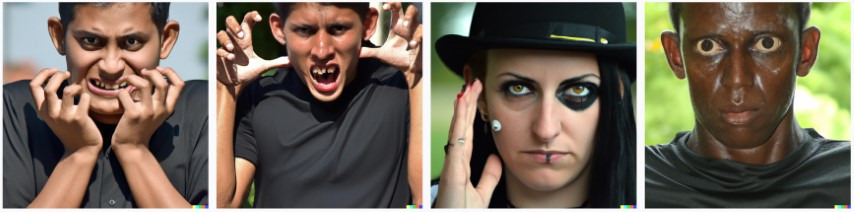}} \hspace{2mm}
\subfloat[Images generated by DALL-E mini with the prompt: \texttt{A very creepooky person}.]{\includegraphics[width=0.45\textwidth]{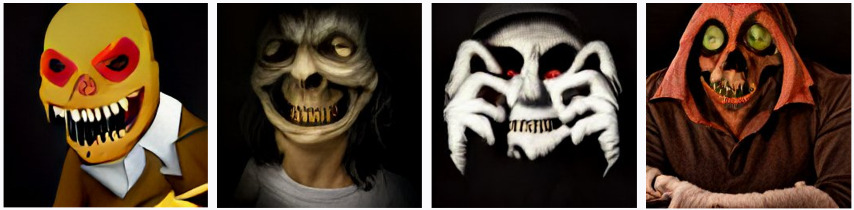}} \hspace{2mm}
\subfloat[Images generated by DALL-E 2 with the prompt: \texttt{A very happeerful person}.]{\includegraphics[width=0.45\textwidth]{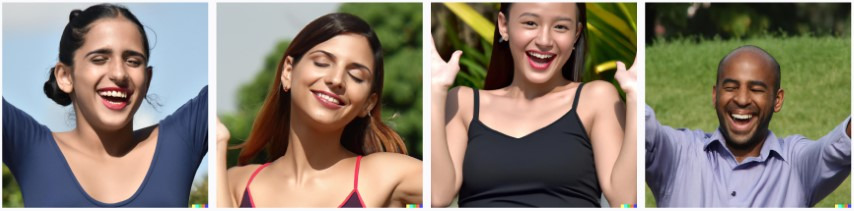}} \hspace{2mm}
\subfloat[Images generated by DALL-E mini with the prompt: \texttt{A very happeerful person}.]{\includegraphics[width=0.45\textwidth]{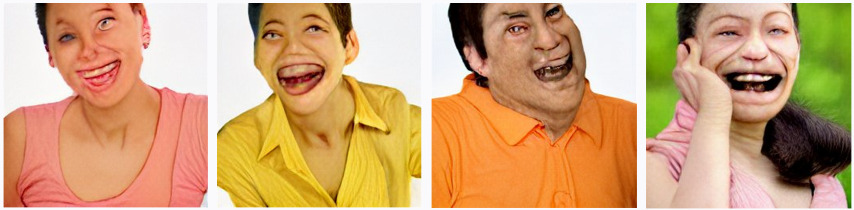}} \hspace{2mm}
\subfloat[Images generated by DALL-E 2 with the prompt: \texttt{A very lovssionate person}.]{\includegraphics[width=0.45\textwidth]{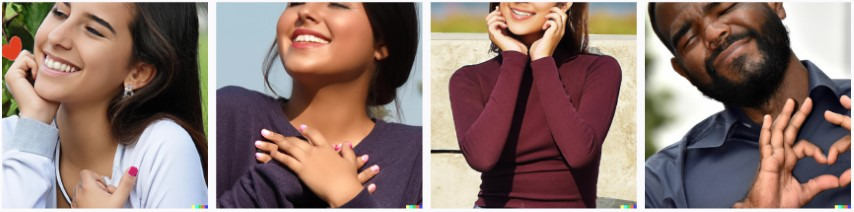}} \hspace{2mm}
\subfloat[Images generated by DALL-E mini with the prompt: \texttt{A very lovssionate person}.]{\includegraphics[width=0.45\textwidth]{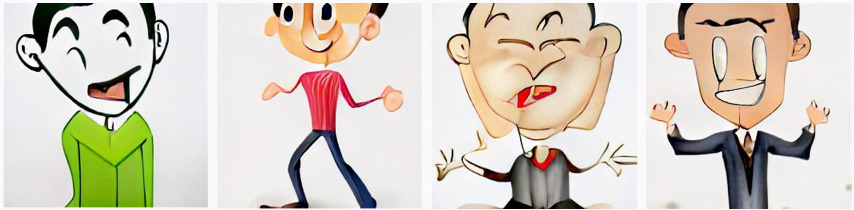}} \hspace{2mm}

\caption{Examples of promtps using monolingual portmanteau words.}
\label{fig:english}
\end{figure}

It is worth mentioning that lexical hybridization can also be effectively applied to prompting within a single language, although this will typically reveal more information about the queried visual concepts, and consequently be less relevant for adversarial attacks (i.e., it is likely that humans could guess the intended meaning of the nonce strings above chance). Fig. \ref{fig:english} shows examples of images generated from multiple models using three invented English portmanteau words: \texttt{creepooky} (\texttt{creepy} + \texttt{spooky}), \texttt{happeerful} (\texttt{happy} + \texttt{cheerful}), and \texttt{lovssionate} (\texttt{loving} + \texttt{compassionate}).

\section{Evocative prompting}

Macaronic prompting relies on existing word chunks to trigger visual associations in text-guided image generation models. By contrast, \emph{evocative prompting} relies more loosely on broad morphological similarity with existing words, word categories, or language-specific features to trigger such associations. The key difference is that nonce strings used for evocative prompting need not include meaningful parts of existing words in any language. Their association with specific visual concepts appears to be mediated instead by the statistical significance of certain letter combinations across words in a given domain.

\begin{figure}[!htp]
\centering 
\subfloat[Images generated by DALL-E 2 with the prompt: \texttt{ceralineus rabaventis}.]{\includegraphics[width=0.45\textwidth]{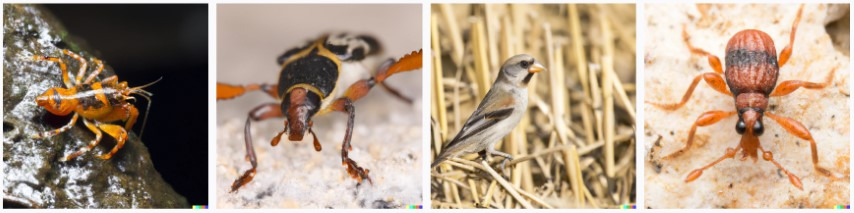}} \hspace{2mm}
\subfloat[Images generated by DALL-E mini with the prompt: \texttt{ceralineus rabaventis}.]{\includegraphics[width=0.45\textwidth]{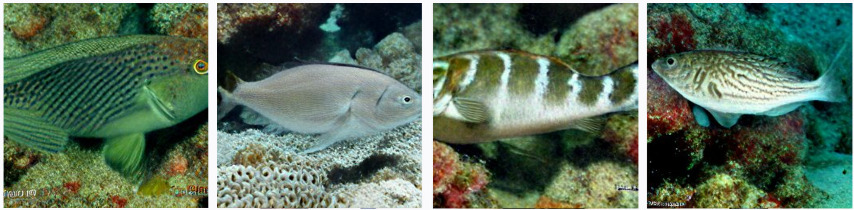}} \hspace{2mm}
\subfloat[Images generated by DALL-E 2 with the prompt: \texttt{rygamera pultris}.]{\includegraphics[width=0.45\textwidth]{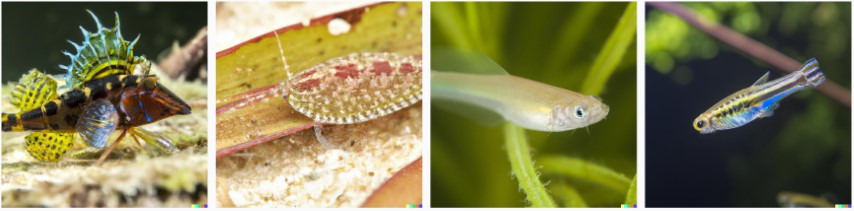}} \hspace{2mm}
\subfloat[Images generated by DALL-E mini with the prompt: \texttt{rygamera pultris}.]{\includegraphics[width=0.45\textwidth]{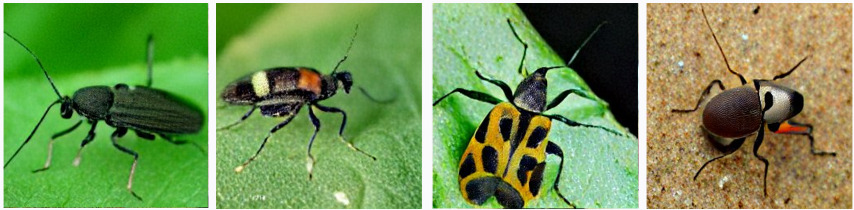}} \hspace{2mm}
\subfloat[Images generated by DALL-E 2 with the prompt: \texttt{bogirus bogirae}.]{\includegraphics[width=0.45\textwidth]{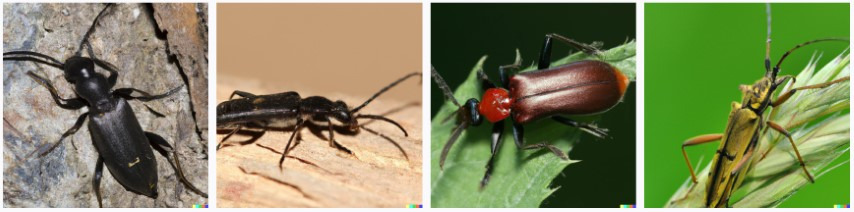}} \hspace{2mm}
\subfloat[Images generated by DALL-E mini with the prompt: \texttt{bogirus bogirae}.]{\includegraphics[width=0.45\textwidth]{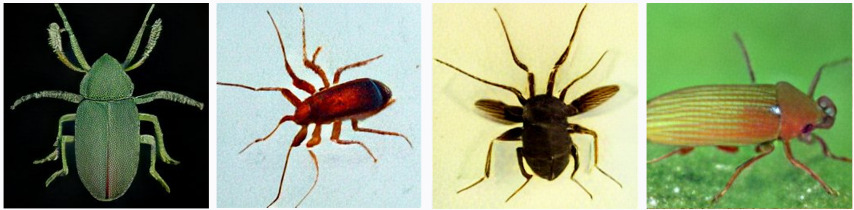}} \hspace{2mm}
\caption{Examples of evocative prompting with pseudolatin nomenclature.}
\label{fig:organisms}
\end{figure}

\begin{figure}[!htp]
\centering 
\subfloat[Images generated by DALL-E 2 with the prompt: \texttt{vacyloraxin}.]{\includegraphics[width=0.45\textwidth]{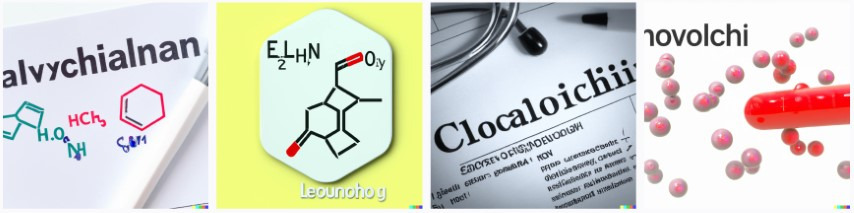}} \hspace{2mm}
\subfloat[Images generated by DALL-E 2 with the prompt: \texttt{walbotricypofen}.]{\includegraphics[width=0.45\textwidth]{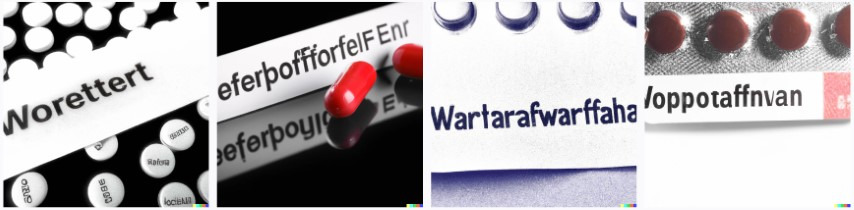}} \hspace{2mm}
\caption{Examples of evocative prompting with fictional medicine names.}
\label{fig:medicine}
\end{figure}

A salient example of evocative prompting concerns the use of pseudolatin that presents superficial similarity to binomial nomenclature for biological taxonomy. This can be done to reliably prompt text-guided image generation models to generate images of the kinds of subjects typically designated through such nomenclature, namely organisms. Fig. \ref{fig:organisms} shows examples of evocative prompting with pseudolatin across models. Another domain in which evocative prompting can easily be achieved pertains to nonce strings that present a superficial similarity to medicine names. Fig. \ref{fig:medicine} displays samples produced with the prompts \texttt{vacyloraxin} and \texttt{walbotricypofen}. 

\begin{figure}[!htp]
\centering 
\subfloat[Images generated by DALL-E 2 with the prompt: \texttt{Woldenbüchel}.]{\includegraphics[width=0.45\textwidth]{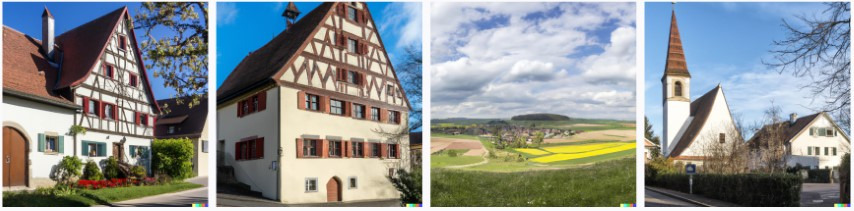}} \hspace{2mm}
\subfloat[Images generated by DALL-E mini with the prompt: \texttt{Woldenbüchel}.]{\includegraphics[width=0.45\textwidth]{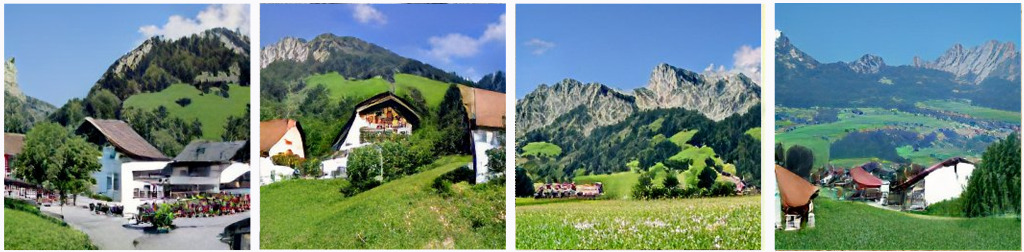}} \hspace{2mm}
\subfloat[Images generated by DALL-E 2 with the prompt: \texttt{Valtorigiano}.]{\includegraphics[width=0.45\textwidth]{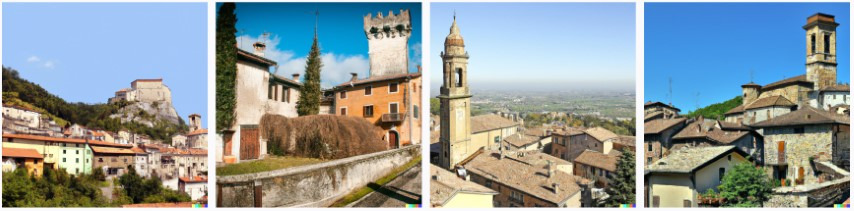}} \hspace{2mm}
\subfloat[Images generated by DALL-E mini with the prompt: \texttt{Valtorigiano}.]{\includegraphics[width=0.45\textwidth]{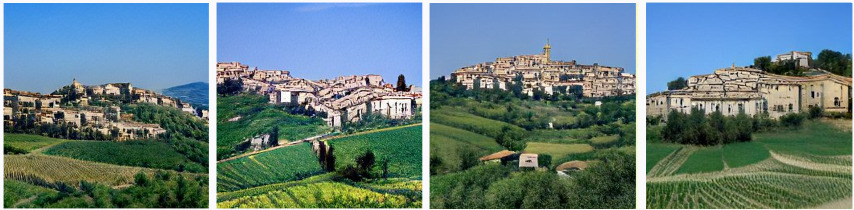}} \hspace{2mm}
\subfloat[Images generated by DALL-E 2 with the prompt: \texttt{Beaussoncour}.]{\includegraphics[width=0.45\textwidth]{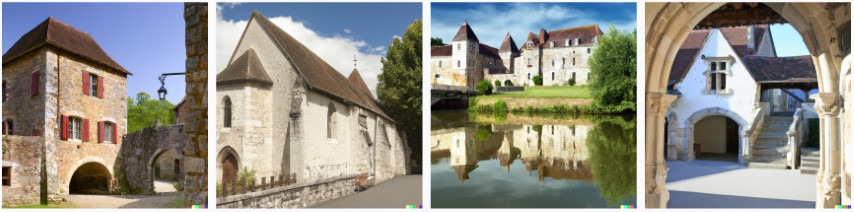}} \hspace{2mm}
\subfloat[Images generated by DALL-E mini with the prompt: \texttt{Beaussoncour}.]{\includegraphics[width=0.45\textwidth]{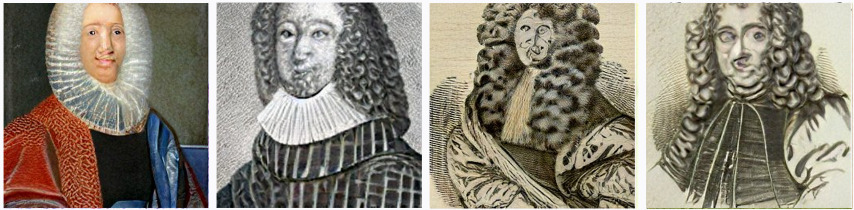}} \hspace{2mm}
\caption{Examples of evocative prompting with geographical associations.}
\label{fig:geography}
\end{figure}

Evocative prompting can also be applied to association of between language-specific features and visual features related to locations and cultures in which the corresponding languages are spoken. Fig. \ref{fig:geography} shows how prompting different models with made-up words that look like names of regional locations in different languages effectively generates images associated with statistically salient features of these locations. Thus, \texttt{Woldenbüchel} generates scenes that look like typical views of a German or Austrian village; \texttt{Valtorigiano} generates scenes that look like typical views of an old Italian town; and \texttt{Beaussoncour} causes DALL-E 2 to generate scenes that look like typical views of a historic French town. Interestingly, this last prompt causes DALL-E mini to generate images reminiscent of 17\textsuperscript{th} century French portraits rather than French locations, but the association with French culture appears to be preserved nonetheless. Furthermore, the images produced by DALL-E mini with the prompts \texttt{voiscellpajaraux} and \texttt{eidelucertlagarzard} in fig. \ref{fig:transfer-failures}c-d could be seen as success cases of evocative prompting (French-looking strings associated with French-looking architecture) instead of failure cases of macaronic prompting. It is intriguing to consider where the boundary lies between macaronic and evocative prompting when real word chunks occur in the prompt; namely, in which cases the model is more sensitive to local features of the prompt (word chunks from any language present in the training data) as opposed to global features (overall morphological similarity to domain-specific words). It seems that differences in training data, model size, and model architecture may cause different models to parse prompts like \texttt{voiscellpajaraux} and \texttt{eidelucertlagarzard} in either ``macaronic'' or ``evocative'' fashion, even when these models are proven to be responsive to both prompting methods.

\begin{figure}[!htp]
\centering 
\subfloat[Images generated by DALL-E 2 with the prompt: \texttt{scariosus ferocianensis}.]{\includegraphics[width=0.45\textwidth]{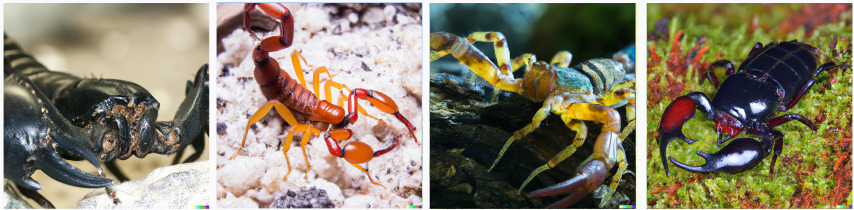}} \hspace{2mm}
\subfloat[Images generated by DALL-E 2 with the prompt: \texttt{cutiosus adorablensis}.]{\includegraphics[width=0.45\textwidth]{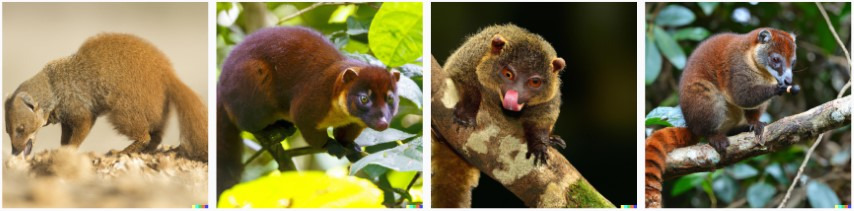}} \hspace{2mm}
\subfloat[Images generated by DALL-E 2 with the prompt: \texttt{watosus swimensis}.]{\includegraphics[width=0.45\textwidth]{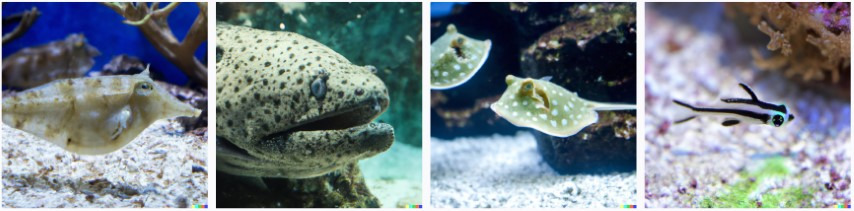}} \hspace{2mm}
\subfloat[Images generated by DALL-E 2 with the prompt: \texttt{flyosus wingensis}.]{\includegraphics[width=0.45\textwidth]{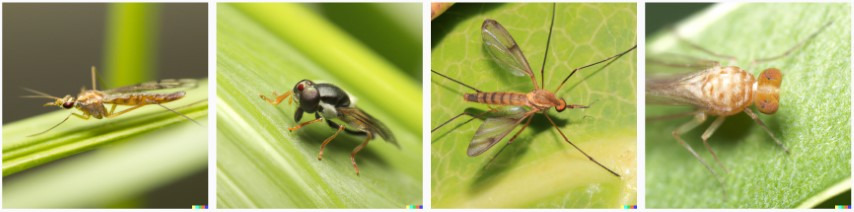}} \hspace{2mm}
\caption{Examples of evocative prompting combined with lexical hybridization.}
\label{fig:english-evocative}
\end{figure}

Evocative prompting can also be combined with lexical hybridization to gain more control over the specific features of the outputs. For example, introducing English word chunks within pseudolatin nomenclature causes DALL-E 2 to generate images of animals with specific properties, as illustrated in fig. \ref{fig:english-evocative}. Thus, the prompt \texttt{scariosus ferocianensis} (combining \texttt{scary} and \texttt{ferocious} with pseudolatin terminations) yields images of conventionally scary ``creepy crawlies'' such as scorpions; \texttt{cutiosus adorablensis} (combining \texttt{cute} and \texttt{adorable} with pseudolatin terminations) yields images of conventionally cute mammals; \texttt{watosus swimensis} (combining \texttt{water} and \texttt{swimming} with pseudolatin terminations) yields images of aquatic animals; and \texttt{flyosus wingensis} (combining \texttt{flying} and \texttt{winged} with pseudolatin terminations) yields images of flying insects.

\section {Discussion}

Vulnerabilty to various kinds of adversarial attacks is a significant limitation of many deep artificial neural networks. First, it highlights ways in which the behavior of these networks may be not be human-like, which has implications for their use as models of human cognition. Second, and more worryingly, adversarial attacks can be intentionally and maliciously deployed to trick neural networks into misclassifying inputs or generating problematic outputs, which may have real-life adverse consequences. 

The new methods presented in this work, \emph{macaronic promping} and \emph{evocative prompting}, enable targeted adversarial attacks on text-guided image generation models. Macaronic prompting involves combining word chunks from different languages into novel hybridized words that lack meaning for human speakers, yet consistently trigger specific visual associations in image generation models. This approach is somewhat related to previous methods using code-mixing for adversarial attacks on multilingual NLP models \citep{tanCodeMixingSesameStreet2021}. The preliminary experiments shown here suggest that hybridized nonce strings can be methodically crafted to generate images of virtually any subject as needed, and even combined together to generate more complex scenes with some success. An obvious concern with this method is the circumvention of content filters based on blacklisted prompts. In principle, macaronic prompting could provide an easy and seemingly reliable way to bypass such filters in order to generate harmful, offensive, illegal, or otherwise sensitive content, including violent, hateful, racist, sexist, or pornographic images, and perhaps images infringing on intellectual property or depicting real individuals. Companies that offer image generation as a service have put a great deal of care into preventing the generation of such outputs in accordance with their content policy. Consequently, macaronic prompting should be systematically investigated as a threat to the safety protocols used for commercial image generation.

Evocative prompting presents a less obvious threat, because it does not offer a principled way of crafting nonce strings that can effectively and reliably trigger specific visual associations. As such, it is mostly limited to vague associations with concepts linked to broad morphological features of words or languages. The present work investigates a few domains in which evocative prompting appears to be somewhat reliable and effective as a proof of concept (biological nomenclature, medicine names, and language-specific geographical/cultural names). It is likely that evocative prompting can be applied to many more subjects whose associated lexicon exhibits salient morphological similarities in various languages. It is worth noting that the combination of evocative prompting with real word chunks can target more specific visual associations (fig. \ref{fig:english-evocative}); however, the resulting prompts are easily interpretable for humans, and may be easier to filter through blacklists.

Overall, maraconic prompting is more viable than evocative prompting as a method for malicious adversarial attacks on text-guided image generation models. It highlights the insufficiency of keyword-based blacklists for content filtration in such models. One way to minimize opportunities for the generation of problematic content is to curate datasets appropriately to exclude specific categories of images or image-caption pairs. For example, some large multimodal datasets on which image generation models are trained are known to contain harmful stereotypes \citep{birhaneMultimodalDatasetsMisogyny2021}. However, dataset curation can only go so far in preventing the downstream generation of problematic content. A well-trained model with the right architecture should in principle be able to construct scenes depicting harmful or offensive imagery by combining individually innocuous concepts present in the training data. A potential complement to prompt filtering through blacklists is the use of a dedicated image classifier to filter visual outputs instead of text inputs. Alternatively, it may be sufficient to feed visual outputs back to the image generation model, retrieve related linguistic concepts in the joint vision-language embedding space, and apply text-based blacklists to the retrieved concepts. Such filtering might prove effective in preventing malicious image generation through text-based adversarial attacks. Another, more radical defense against such attacks would involve filtering all prompts whose words do not occur in a comprehensive multilingual vocabulary list, which would effectively prevent the use of nonce strings across the board. However, this would impose significant limitations of the use of image generation models, particularly when it comes to prompts containing proper names or resilience to typographical errors.

Unlike adversarial perturbations traditionally used in computer vision, attacks from macaronic or evocative prompting are not completely inscrutable to humans \cite{bucknerUnderstandingAdversarialExamples2020}. With enough knowledge of languages and morphological features, it should be possible to recover the intended visual associations of nonce strings used in these approaches. While this work has only investigated macaronic hybridized strings using five languages (English, French, German, Italian, and Spanish), it should be possible to make these nonce strings even more cryptic by adding word chunks from additional lesser-known languages, especially as image generation models get trained on increasingly multilingual datasets. There is another respect in which these techniques can remain somewhat inscrutable, in that it seems currently difficult to predict how any specific model will respond to a given macaronic or evocative prompt. This is apparent, for example, from the fact that DALL-E mini is vulnerable to some but not all macaronic prompts that prove effective with DALL-E 2 (fig. \ref{fig:transfer-failures}), or that it has different visual associations with some but not all evocative prompts (fig. \ref{fig:geography}). It is currently challenging to disentangle the factors that determine convergent and divergent behavior with macaronic and evocative prompting across models.

\section{Conclusion}

Text-guided image generation models are not immune to adversarial attacks. While various properties of these models -- including size, architecture, tokenization prodecure and training data -- may influence their vulnerability to text-based adversarial attacks, preliminary evidence discussed in this work suggests that some of these attacks may nonetheless work somewhat reliably across models. Further research is needed to determine the factors that determine how different models respond to macaronic and evocative prompting, as well as effective strategies to mitigate malicious use of these techniques for the generation of harmful, offensive, or otherwise problematic visual content.

\end{document}